\documentclass{article}
\usepackage[a4paper]{geometry}
\usepackage{listings}
\usepackage{booktabs} 
\usepackage{multicol}
\usepackage{bera}
\usepackage{caption}
\usepackage{chemformula}
\usepackage{PRIMEarxiv}
\usepackage[utf8]{inputenc} 
\usepackage[T1]{fontenc}
\usepackage{hyperref}       
\usepackage{url}            
\usepackage{booktabs}       
\usepackage{amsfonts}       
\usepackage{nicefrac}       
\usepackage{microtype}      
\usepackage{lipsum}
\usepackage{xcolor}
\usepackage{wrapfig}
\usepackage{tcolorbox}
\usepackage{adjustbox}
\usepackage{amsmath}
\usepackage{tikz}
\usepackage{fancyhdr}       
\usepackage{graphicx}       
\graphicspath{{media/}}     
\usepackage[inkscapeformat=png]{svg}
\usepackage[T1]{fontenc}
\usepackage[utf8]{inputenc}

\newcommand{\source}[1]{\caption*{Source: {#1}} }

\usepackage[backend=biber, style=numeric, sorting=none]{biblatex}
\title{Efficient Language Adaptive Pre-training: Extending State-of-the-Art Large Language Models for Polish}
\author{
  Szymon Ruciński\\
  \textit{Apostroph Group - Artificial Intelligence Laboratory} \\ \\
  Zürich, Switzerland \\
  \texttt{\{Szymon Ruciński\}@apostrophgroup.ch} \\
}

\begin{document}
\maketitle

\begin{abstract}
This study explores the potential of fine-tuning foundational English Large Language Models (LLMs) for generating Polish text. The first step involves Language Adaptive Pre-training (LAPT) on a high-quality dataset of 3.11 GB, consisting of 276 million Polish tokens. The LAPT is followed by additional fine-tuning aimed at solving nine  \href{https://klejbenchmark.com}{KLEJ challenges} \cite{rybak2020klej}. Our trained model \href{https://huggingface.co/szymonrucinski/Curie-7B-v1}{Curie-7B-v1} not only generates Polish text with the lowest perplexity of 3.02 among decoder-based Polish models but also closely rivals the performance of the best Polish encoder-decoder models with a less than 2\% gap on 8 out of 9 tasks. Curie-7B-v1 used approximately 2-3\% of a typical dataset size to learn Polish. The LAPT was completed in less than five days using a consumer GPU, highlighting the method’s efficiency.
\par
The proficiency of the model in Polish was significantly enhanced, demonstrating the viability of this approach for adding new languages to existing LLMs by training just 1.2\% of its parameters. To contribute to the community's collaborative progress, the model has been released as open-source.
\end{abstract}


\section{Introduction}
LLMs have enhanced the efficiency of many natural language processing (NLP) tasks. This improvement comes with the trade-off of resource-intensive pre-training and inference. At the pre-training phase model gains a general understanding of language, including grammar rules, linguistic patterns, factual information, and reasoning abilities \cite{bommasani2022opportunities}. Currently, all of the \href{https://huggingface.co/spaces/HuggingFaceH4/open_llm_leaderboard}{best open-source LLMs} are pre-trained on mostly English data. As per the findings of Web Technology Surveys3 \cite{w3techs2024contentlanguage}, more than 51.7\% of the content on the internet is in English, while data in over 100 non-English languages accounts for just 48.3\% of the total. The Polish language contributes to just 1.6\% of the Internet's content. Due to data insufficiency, it is significantly harder to develop a non-English LLM.
\par
The performance of LLMs is influenced by several crucial factors, including the number of model parameters, the number of observed tokens, and the overall quality of the text \cite{xue2023repeat} \cite{DBLP:journals/corr/abs-2001-08361}.
Ideally, the pre-training dataset should scale with the number of model parameters \cite{xue2023repeat}. The resource-intensive nature of pre-training LLMs poses a challenge for low-resource languages such as Polish. For comparison, Meta's LLama 2 was trained on 2 trillion tokens \cite{touvron2023llama} and GPT-3 on roughly 300 billion tokens \cite{brown2020language}. As of today, to the best of the author's knowledge, there are no high-quality open-source datasets of Polish text exceeding 100 billion tokens in size. Developing LLM is a substantial investment. For the sake of comparison, it is claimed that GPT-4 cost is over \$100,000,000,  MistralAi's Mistral-7B cost \$500,000 to train,  Meta's LLaMa2 70b was trained on 2048 A100 GPUs for 23 days which is estimated to cost around \$2,000,000. These are the costs of just a plain LLM pre-training without including the costs of e.g. data collection or human evaluation necessary to turn these into complex AI assistants or classifiers.
\par
 Pre-training isn't the only technique to adopt LLMs to low-resource languages. This can also be done via transfer learning \cite{zhuang2020comprehensive} \cite{DBLP:journals/corr/abs-2004-10964}, fine-tuning LLM for Causal Language Modeling (predicting the next element in a sequence iteratively) \cite{wu2023metalearning} in a supervised manner on text in a language it has merely or never seen in a pre-training phase. LAPT for text generation in a specific language, such as Polish is a potentially effective strategy. For instance, studies have shown that Domain Adaptive Pre-training can significantly improve the performance of foundational LLMs in clinical tasks \cite{gema2023parameterefficient} \cite{karn-etal-2023-shs} \cite{chen2023meditron70b}. 
LLaMA \cite{touvron2023llama}, when equipped with a LoRA adapter fine-tuned on medical texts, particularly outperforms foundational models in clinical domain tasks \cite{gema2023parameterefficient}. The study \cite{gema2023parameterefficient} demonstrates that this approach yields substantial improvements, especially in large-scale multilabel classification tasks like diagnoses and procedures classification. This marks a significant advancement over existing custom-trained language models, highlighting the efficiency of LoRA  Domain Adaptive Pre-training in highly specialized domains.

While the specific application to Polish wasn't addressed in the papers we found, the principles of Domain Adaptive Pre-training are widely applicable across languages. This is especially relevant in the context of neural machine translation \cite{bapna2019simple} and cross-lingual tasks, where models are often adapted to new languages and domains to improve their performance. This could be directly applicable to a language like Polish by enabling the model to better capture the syntax, semantics, and unique idiomatic expressions, leading to more accurate and contextually appropriate text generation.
The preliminary evaluations \ref{tab:test-mistral-lang-understanding} have revealed that Mistral-7B English open-source LLM exhibits a basic ability to generate and understand texts in Polish. This capability could be leveraged to significantly improve Polish text generation and comprehension.
\begin{table}[!ht]
\begin{center}
\setlength{\abovecaptionskip}{4pt} 
\begin{tabular}{c c}
\toprule
\textbf{Input Tokens} & \textbf{Generated Tokens} \\
\midrule
Szybkie czerwone autko & jeździące po coś w tym kraju \\
Kanapka & ze kokosovým mlékem je moje oblíbená receptura \\
Król Karol & wręczył Trzeciej Wikitorii Królowej \\
Sport to zdrowie & i dlatego zauważyć, ze zdrowa dieta \\
\bottomrule
\end{tabular}
    \caption{Example of Token Sequences Generated by Mistral-7B.}
    \label{tab:test-mistral-lang-understanding}
\end{center}
\end{table}
The applications of LLMs include diverse domains such as online retail, medicine, education, engineering, linguistics, and the gaming industry. The integration of these AI solutions not only enhances business productivity but also yields substantial economic benefits. The introduction of LLM-based AI tools has demonstrated a marked improvement in operational efficiency, evidenced by a 14\% average increase in the rate of completed tasks per hour \cite{brynjolfsson2023generative}.
In the context of Poland, the usage of such technologies is almost instant but the development and research of custom solutions is progressing at a slower pace. As of now, there isn't a cutting-edge, high-quality LLM designed exclusively for Polish.  This delay forces businesses to depend on externally hosted solutions, such as OpenAI's ChatGPT, particularly in the realm of digital assistants. While these external solutions offer immediate benefits, they also entail financial costs and limit control over data flow.  The reliance on external AI technologies, while a temporary solution, underscores the need for the development and deployment of localized LLMs to ensure data sovereignty and capitalize on the economic and technological potential of AI.
\par
This study aims to ascertain whether utilizing an established LLM solution can facilitate the creation of versatile Polish-adapted LLM that is both time-efficient and economically viable. This approach, which involves further building a classifier/regressor on top of LAPT model fine-tuned to solve a domain-specific downstream task that is applicable for business use cases (sentiment analysis, predicting/labelling online reviews, generating texts). 

\par The following \textbf{Research Question (RQ)} have been defined and will be addressed in this paper:
\begin{table}[ht]
\centering
\begin{tabular}{p{0.05\linewidth} p{0.7\linewidth}}
    \textbf{RQ 1} & How well does our model Curie-7B-v1 generate Polish text? \\
    \textbf{RQ 2} & How does LAPT LLM perform against top models in KLEJ benchmark? \\
    \textbf{RQ 3} & What are the estimated costs, time requirements, and energy consumption involved in building a model like Curie-7B-v1? \\
\end{tabular}
\end{table}

\clearpage




\section{Methodology}
\label{sec:headings}
This subsection provides a clear overview of the experiments conducted in this study. It explains the steps followed and the techniques used. The mathematical principles underlying these experiments are also described. Furthermore, this section discusses the metrics used to evaluate the results of the experiments. 
\subsection{Language Adaptive Pre-training}
Given a pre-trained LLM $P_{\Phi}(y|x)$ its parameters $\Phi$ and a training dataset $\mathcal{Z} = \{(x_i, y_i)\}_{i=1,...,N}$. In order to adapt to the new language, the model weights need be to updated iteratively from its pre-trained state $\Phi_0$ to $\Phi = \Phi_0 + \Delta\Phi$. The process of maximising the objective function can be defined as follows:
\par
\begin{equation}
\underset{\Phi}{\mathrm{argmax}} \sum_{(x,y) \in \mathcal{Z} } \sum_{t=1}^{|y|} \log (P_{\Phi}(y_t | x, y_{<t}))
\label{eq:classical_finetuning}
\end{equation}
\par
This task is computationally intensive and demands substantial resources. In the classical paradigm \ref{eq:classical_finetuning}, a full fine-tuning means that the model needs to learn a \(\Delta \Phi\) whose dimension is equal to the entire pre-trained parameters \(|\Delta \Phi| = |\Phi_0|\), which is computationally expensive.
In the proposed paradigm (\ref{eq:peft_finetuning}) LoRA (Low-Rank Adaptation) \cite{hu2021lora} used in this study we tune only small additional parameters \(\theta\) such that \(\Phi = \Phi_0 + \Delta \Phi(\theta)\). Its dimension is very small compared to the original parameters \(|\theta| \ll |\Phi_0|\). Thus, the training can be expressed as:
\begin{equation}
\underset{\theta}{\mathrm{argmax}} \sum_{(x,y) \in \mathcal{Z}} \sum_{t=1}^{|y|} \log (P_{\Phi+\Delta\Phi(\theta)} (y_t | x, y_{<t}))
\label{eq:peft_finetuning}
\end{equation}

In the classical paradigm (\ref{eq:classical_finetuning}), the outcome of LAPT would be a Polish-adapted LLM. While in used paradigm (\ref{eq:peft_finetuning}), the outcome would be the Polish LoRA adapter \cite{DBLP:journals/corr/abs-2106-09685}, which can be combined with the untouched foundational LLM to generate Polish text.
\begin{figure}[h]
\begin{center}
\center
\includegraphics[width=12cm]{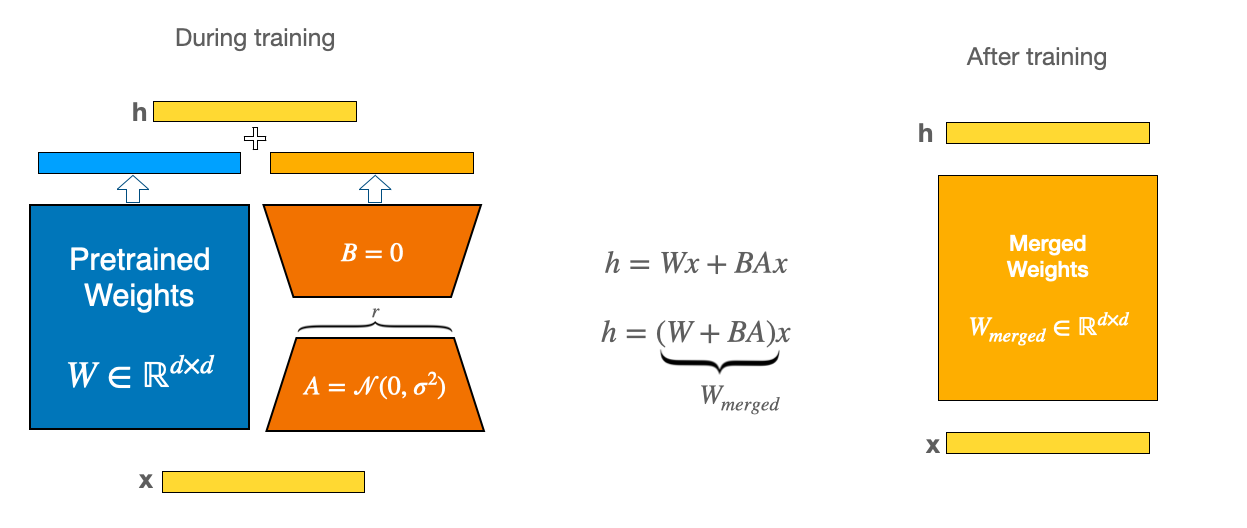}
    \caption{LoRA Diagram.} \label{imageLabel}
    \source{\href{https://huggingface.co/docs/peft/conceptual_guides/lora}{Huggingface.com}}
\end{center}
\setlength{\abovecaptionskip}{4pt}
\end{figure}
\par
Perplexity is a commonly used metric in natural language processing (NLP) to evaluate the quality of LLMs. In the context of text generation, perplexity indicates how well the language model predicts the sequence of words in a given test text. It's a measure of how "surprised" the model is by the data it's seeing. A lower perplexity score indicates that the model is better at predicting the sample.
It is defined as follows let \(P_i\) be the perplexity of the \(i\)-th sentence in the batch, calculated as: \(P_i = 2^{H_i}\) where  \(H_i\) is the average cross-entropy for the \(i\)-th sentence, given by:
\begin{equation}
H_i = -\frac{1}{N_i} \sum_{j=1}^{N_i} \log_2 (P(w_{ij} | w_{i1}, w_{i2}, \ldots, w_{ij-1}))
\end{equation}
Here,  \(N_i\) is the number of words in the \(i\)-th sentence, and
\(P(w_{ij} | w_{i1}, w_{i2}, \ldots, w_{ij-1})\) is the predicted probability of the \(j\)-th word given the preceding context within the sentence. Then, the mean perplexity across the batch of \(M\) sentences is defined as follows: 
\begin{equation}
\overline{P} = \frac{1}{M} \sum_{i=1}^{M} P_i
\end{equation}
\par
\subsection{Fine-tuning For Downstream Tasks}
After a language model is fine-tuned according to (\ref{eq:classical_finetuning}) it needs to solve a downstream task, such as sentimental analysis of online reviews. A pre-trained LLM \(P_{\phi,\Theta}\) with its domain-adapted parameters \(\Phi\) and a newly initialised classification layer \(\Theta\), as well as a training dataset \(\mathcal{Z} = \{(x_i, y_i)\}_{i=1,...,N}\) has a task to minimize a specific loss function, such as a cross-entropy loss \cite{gema2023parameterefficient}:
\begin{equation}
\underset{\Phi,\Theta}{\mathrm{argmax}}{\frac{1}{N} \sum_{i=1}^{N} y_i \log (P_{\phi,\Theta} (x_i))}
\end{equation}

In the proposed paradigm (\ref{eq:peft_finetuning}), the fine-tuning process only updates the small additional parameters \(\Delta\Phi(\theta)\) and the classifier head \(\Theta\) :
\begin{equation}
\underset{\theta,\Theta}{\mathrm{argmax}} \frac{1}{N} \sum_{i=1}^{N} y_i \log (P_{\phi+\Delta\Phi(\theta),\Theta} (x_i))
\end{equation}
\par
\subsection{Data}
This subsection details the datasets used in both the LAPT phase and the second phase for addressing downstream tasks.
\subsubsection{The Dataset for Language Adaptive Pre-training}
For the initial phase of the LAPT the SpeakLeash \cite{SpeakLeash2024} dataset was used. It offers an extensive and diverse collection of texts in Polish. It consists of 1TB of a wide range of texts in Polish. Only the highest quality extract of approximately 2 GB was selected.
\begin{figure}[!h]
\begin{center}
\centering
\includegraphics[scale=0.7]{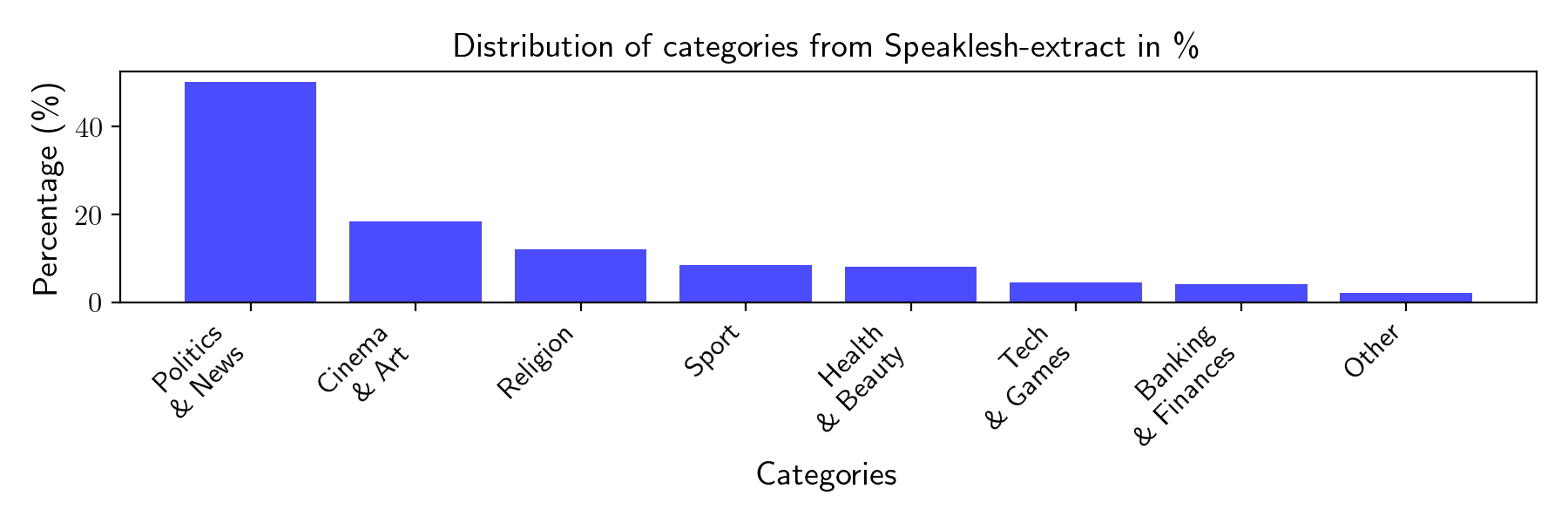}
\caption{Distribution of Categories Used in the LAPT Phase.}
\end{center}
\setlength{\abovecaptionskip}{4pt}
\label{fig:categories-distribution}
\end{figure}


\begin{tcolorbox}[colback=white,colframe=black!50, rounded corners, title=Sample Extracts]
"Z Podwala Staromiejskiego zniknęły stragany z warzywami i owocami ..."
\tcblower
"Transmisja cyfrowych danych w sieciach GSM Tworzenie standardu GSM rozpoczęło się w 1982 roku, kiedy to powołano do działalności zespół roboczy, przed którym postawiono zadanie opracowania założeń ..."
\end{tcolorbox}

Speaklesh dataset has been curated to include texts from a variety of sources, ensuring a comprehensive representation of the Polish language. At LAPT we specifically trained adapter on online texts sourced from hundreds of Polish web portals along with extensive extract from Polish Wikipedia. This approach was instrumental in covering a broad spectrum of topics and writing styles, thus enhancing the adaptability and accuracy of our model. The merged dataset consisted of 2,157,867 texts.

\par
\subsubsection{Downstream Tasks} \label{klej-tasks}
The \href{https://klejbenchmark.com/}{KLEJ Benchmark}\cite{rybak2020klej} consists of 9 tasks (\ref{tab:KLEJ-datasets}) for evaluating the performance of language models. Each task is designed to assess different aspects of language processing abilities, such as understanding context, recognizing emotions, and identifying specific entities in text. This benchmark provides a comprehensive framework for testing and comparing the capabilities of various language models in processing and understanding the Polish language.

\begin{multicols}{2}
    \begin{itemize}
        \item \textbf{NKJP-NER:} Predict the type of a named entity in sentences from the NKJP corpus. \\
        \item \textbf{CDSC-E:} Determine entailment between pairs of sentences from the Compositional Distributional Semantics Corpus. \\
        \item \textbf{CDSC-R:} Assess semantic relatedness between sentence pairs in the Compositional Distributional Semantics Corpus. \\
        \item \textbf{CBD:} Detect cyberbullying content in Twitter messages from the 2019 PolEval competition. \\
        \item \textbf{PolEmo2.0-IN:} Predict the sentiment of online reviews in medicine and hotel domains. \\
    \end{itemize}
    
    \columnbreak
    
    \begin{itemize}
        \item \textbf{PolEmo2.0-OUT:} Predict sentiment for out-of-domain reviews like products and university. \\
        \item \textbf{DYK:} Decide if an answer to a question is correct in the 'Did You Know' dataset. \\
        \item \textbf{PSC:} Identify summaries of the same or different news articles in the Polish Summaries Corpus. \\
        \item \textbf{AR:} Predict product ratings from 1 to 5 in the Allegro Reviews dataset. \\
    \end{itemize}
\end{multicols}

\begin{table}[htbp]
\centering
\begin{tabular}{lllllll}
\toprule
\textbf{Task-Name} & \textbf{Train} & \textbf{Dev} & \textbf{Test} & \textbf{Domain} & \textbf{Metrics} & \textbf{Objective} \\
\hline
\multicolumn{7}{c}{Single-Sentence Tasks} \\
\hline
NKJP-NER & 16k & 2k & 2k & Balanced corpus & Accuracy & NER classification \\
CDSC-R & 8k & 1k & 1k & Image captions & Spearman corr. & Semantic relatedness \\
CDSC-E & 8k & 1k & 1k & Image captions & Accuracy & Textual entailment \\
\hline
\multicolumn{7}{c}{Multi-Sentence Tasks} \\
\hline
CBD & 10k & - & 1k & Social Media & F1-Score & Cyberbullying detection \\
PolEmo2.0-IN & 6k & 0.7k & 0.7k & Online reviews & Accuracy & Sentiment analysis \\
PolEmo2.0-OUT & 6k & 0.5k & 0.5k & Online reviews & Accuracy & Sentiment analysis \\
Czy wiesz? & 5k & - & 1k & Wikipedia & F1-Score & Question answering \\
PSC & 4k & - & 1k & News articles & F1-Score & Paraphrase \\
AR & 10k & 1k & 1k & Online reviews & wMAE & Sentiment analysis \\
\hline
\end{tabular}
\setlength{\abovecaptionskip}{4pt}
\caption{KLEJ Datasets and Their Characteristics.}
\label{tab:KLEJ-datasets}
\end{table}



\subsection{Model Selection}
In the research of an optimal base foundational LLM for the LAPT to build Curie-7B-v1, we have identified Mistral-7B, developed by the French startup \href{https://mistral.ai/}{Mistral}, as the suitable foundational model. Among the open-source models evaluated, which include LLama2, Falcon, and Bloom, Mistral-7B demonstrates elementary proficiency in processing and interpreting the Polish language. This proficiency is a decisive factor, given the LAPT's primary focus on the Polish language. Furthermore, Mistral-7B distinguishes itself also through several key features beyond its language capabilities:
\begin{itemize}
    \item \textbf{Performance}: Mistral 7B shows exceptional performance, consistently outperforming Llama 2 13B and competing effectively with Llama 30B in various tasks.
    \item \textbf{Architectural Advancements}:
    \begin{itemize}
        \item \textbf{Grouped-Query Attention}: Enhances processing efficiency, leading to faster inference times.
        \item \textbf{Sliding-Window Attention}: Improves handling of longer data sequences while maintaining computational efficiency.
        \item \textbf{Byte-fallback BPE Tokenizer}: Ensures effective management of a broad spectrum of textual inputs.
    \end{itemize}
    \item \textbf{Context Window}: The model's ability to refer to a significant amount of previous information enhances its performance in continuous tasks. The model has a window size of 4096 and the context length is 8192. It is significantly more than any vanilla BERT \cite{DBLP:journals/corr/abs-1810-04805} model could handle.
    
    \item \textbf{Accessibility}: In \textit{float32} precision, Mistral 7B requires $\sim28$GB of VRAM, while in \textit{float16} precision, it needs $\sim14$GB. This makes it accessible for consumer-grade GPUs.
    
    \item \textbf{Versatility}: Mistral 7B excels in English language processing and coding tasks, making it versatile for various enterprise applications.
    
    \item \textbf{Open-Source License}: Available under the Apache 2.0 license, it encourages community-driven development and transparency.
\end{itemize}

\section{Experiments}

\subsection{Hardware and Software Stack}
For the experiments, a server from a leading cloud provider was leased, featuring formidable computational resources. This included the \href{https://www.nvidia.com/content/dam/en-zz/Solutions/design-visualization/rtx-6000/proviz-print-rtx6000-datasheet-web-2504660.pdf}{NVIDIA RTX A6000 ADA GPU} with 48GB of VRAM, paired with the powerful AMD Epyc 7742 processor, which has 64 cores and 128 GB of RAM and 1TB SSD M2 drive. The server ran on a Linux Ubuntu platform, equipped with a Conda environment with Pytorch 2.0 and the latest 12.2 CUDA driver.
\subsection{The Adaptive Pre-training}
\label{adaptive-pretraining-experiment}
\begin{wraptable}{R}{5.5cm}
\vspace{-15pt}
\begin{center}
\setlength{\abovecaptionskip}{4pt} 
\setlength{\belowcaptionskip}{4pt} 
\begin{tabular}{c c}
\toprule
\textbf{Hyperparameters} & \textbf{Value} \\
\midrule
lora\_rank & 32 \\
lora\_dropout & 0.05 \\
lora\_alpha & 16 \\
warmup\_steps & 0.1 \\
learning\_rate & \(2.5 \times 10^{-5}\) \\
neftune\_noise\_alpha & 2 \\
batch\_size & 128 \\
max\_seq\_len & 128 \\
\bottomrule
\end{tabular}
\caption{Used Hyperparameters.}
\label{tab:used-hyperparams}
\end{center}
\vspace{-20pt}
\end{wraptable}
The model underwent training utilizing the AdamW optimizer \cite{DBLP:journals/corr/abs-1711-05101}, with a learning rate (LR) schedule featuring an initial warm-up at a rate of 0.1, followed by a reduction in the final LR to 10\% of its peak value. The LoRA adapter was configured with standard settings, including a rank of 32 and $\alpha$ value of 16, complemented by a LoRa dropout rate of 0.05. 
To enhance the performance NEFTune noise was added to the embedding vectors during the training process. \cite{jain2023neftune}. The maximum input size was set to 128 tokens, with a batch size of 128.
\par
\begin{figure}[!h]
\begin{center}
\centering
\includegraphics[scale=0.9]{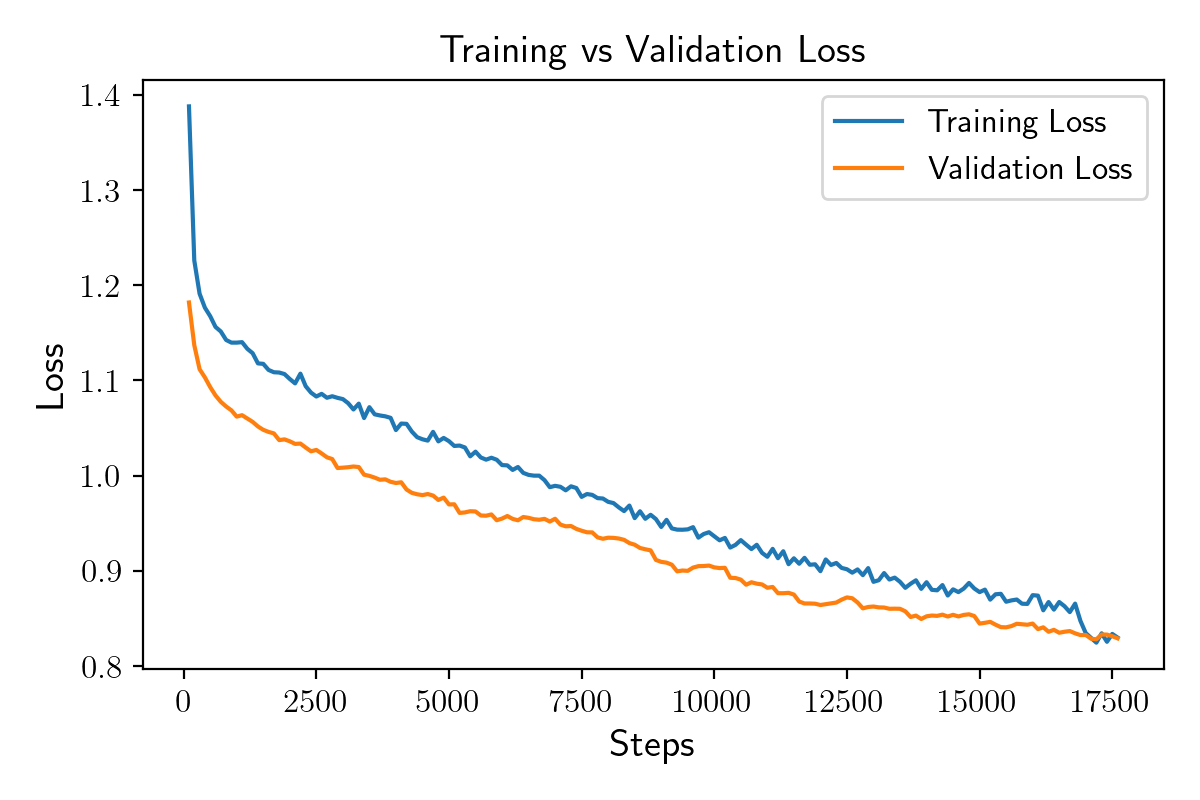}
\caption{Training progess.}
\label{fig:training-progress}
\end{center}
\end{figure}

 
Training of the final model was completed in exactly one epoch, requiring a total of 106 hours. The training was not extended beyond this duration due to the onset of overfitting after the first epoch, which led to the generation of nonsensical text. Figure \ref{fig:training-progress} illustrates the model's training loss and validation. Loss and performance evaluations indicated that optimal model quality was attained after the initial epoch, with further training resulting in a significant decline in quality, ultimately producing incoherent text. The model was validated using 1000 distinct examples coming from the training set. 
\clearpage
\textbf{RQ 1: How well does our model Curie-7B-v1 generate Polish text?}\label{answer-rq1}\\ 
To evaluate the model's performance, the perplexity scores were compared before and after fine-tuning. This comparison serves as an objective measure to assess the effectiveness of the LAPT process.
Low perplexity calculated on the test set of 1000 (never seen before) examples prove that the model has a significantly better understanding of the Polish language after fine-tuning.  Results signify that adapting the LLM to the new language has been successful. 
\begin{table}[h!]
\centering
\begin{tabular}{cccccc}
\toprule
Model-Name & Average Perplexity \(\downarrow\) & Size (Billion Parameters) & Modality & Main Language & Tokens seen \\
\midrule
\textbf{Curie-7B-v1} & \textbf{3.02} & 7.24 & Pre-trained & Polish & \(\ast276\) Million \\
\textbf{Mistral-7B-v1} & 6.82 & 7.24 & Fine-tuned & English & Unknown \\
LLama2-7B & 7.71 & 6.74 & Pre-trained & English & 2 Trillion \\
APT3-1B-Base & 23.30 & 1.04 & Pre-trained & Polish & 60 Billion\\
Polish-GPT2-XL & 97.37 & 1.67 & Pre-trained & Polish & Unknown \\
\bottomrule
\end{tabular}
\setlength{\abovecaptionskip}{4pt}
\caption{Average Perplexity of Models with Additional Information.}
\footnotesize{*Model was fine-tuned using 276 million Polish tokens, the initial count of tokens it was pre-trained on is not included.}
\label{table:extended}
\end{table}
The LAPT model Curie-7B was compared against the excellent English LLMs and the two well-established Polish decoder-only models \href{https://huggingface.co/sdadas/polish-gpt2-xl}{Pol-GPT-2} and \href{https://huggingface.co/Azurro/APT3-1B-Base}{APT3-1B-Base} (based on LLAMA architecture). Our solution surpassed all the others by a notable margin. Additional empirical evaluations indicate that the adapted model demonstrates a high degree of linguistic competence, as reflected by its capacity to generate coherent and contextually relevant text. Most significantly, the model achieves the lowest perplexity score when benchmarked against other language models. 
\subsection{Fine-tuning for KLEJ downstream tasks}
The model, an outcome of the experiments detailed in (\ref{adaptive-pretraining-experiment}), served as the foundation for developing classifiers and regressors to address the KLEJ tasks (\ref{klej-tasks}). A prevalent issue in the datasets was the strong class imbalance, which was mitigated using weighted cross-entropy. The training duration for classifiers ranged between 2 to 4 hours on average. These classifiers underwent training for 20 epochs, incorporating an early stopping parameter set at 5. Hyperparameter tuning was employed for optimizing parameters. Minimal to no data preprocessing was applied. In instances lacking a validation dataset, a stratified 20\% segment of the training dataset was utilized as a control sample.

\textbf{RQ 2: How does LAPT LLM perform against top models in KLEJ benchmark?} \label{answer-rq2}\\
Our model Curie-7B-v1 a decoder-only model fine-tuned on 276 million tokens handled 8 challenges exceptionally well. It was extremely close to the best baseline model which is a native Polish model trained on significantly more Polish data. LAPT used the least amount of tokens and yet the model was powerful enough to obtain results comparable with the current SOTA in the 8 out of 9 tasks.
Curie-7B-v1 used significantly less data and in 8 tasks got the average of 89.35 using just based on an estimation between 2-3\% of the dataset size of the best model that scored 90.7\% in those tasks.
Although our model is bigger than the Polish RoBERTa-v2 (large) it requires significantly fewer tokens to learn a new language, Polish.

\begin{table}[htbp]
\centering
\begin{adjustbox}{width=1\textwidth, center=\textwidth}
\begin{tabular}{@{}ccccccccccccc@{}}
\toprule
& Model-Name                           & NKJP-NER & CDSC-E & CDSC-R & CBD & PolEmo2.0-IN & PolEmo2.0-OUT & DYK & PSC & AR \\
\midrule
& \textbf{Curie-7B-v1}             & 93.4     & 92.2   & 94.9   & 49.0 & 92.7         & 80.0          & 76.2 & 98.6 & 86.8 \\
& Polish RoBERTa-v2 (large)        & 95.8     & 94.3   & \textbf{95.1}   & \textbf{74.3} & \textbf{93.1}         & \textbf{84.0}          & 75.4 & 98.8 & \textbf{89.2} \\
& HerBERT (large)                  & \textbf{96.4}     & 94.1   & 94.9   & 72.0 & 92.2         & 81.8          & 75.8 & 98.6 & 89.1\\
& XLM-RoBERTa (large) + NKJP       & 94.2     & \textbf{94.2}   & 94.5   & 72.4 & \textbf{93.1}         & 77.9          & \textbf{77.5} & \textbf{98.9} &  88.2\\
& Polish RoBERTa (large)           & 94.5     & 93.3   & 94.9   & 71.1 & 92.8         & 82.4          & 73.4 & 98.8 & 88.8 \\
\bottomrule
\end{tabular}
\end{adjustbox}
\setlength{\abovecaptionskip}{4pt}
\caption{Models Comparison.}
\label{table:model_comparison}
\end{table}

The model shows low performance in the cyber-bullying detection (CBD) task. This underperformance is attributed to the model's lack of exposure to a wide range of swear words. Additionally, the ambiguity of some insults, which can have double meanings, confuses the model. The dataset employed was primarily composed of news articles, literature, and texts, which utilize formal or semi-formal language and exclude inappropriate phrases.
LAPT used the least amount of tokens when compared to baselines.  However, this was enough to obtain results almost on pair with the current SOTA. 
\begin{table}[htbp]
\centering
\begin{tabular}{@{}ccccc@{}}
\toprule
Model-Name & Batch Size & Update Steps & Corpus Size & Tokens Seen \\
\midrule
Curie-7B & 128 & 17k & ~3.11GB & *276 Million \\
Polish RoBERTa-v2 (large) & 2k & 400k & ~200GB & **15-30 Billion \\
Herbert (large) & 2.5k & 60k & Unknown & ~8.6 Billion \\
XLM-RoBERTa (large) + NKJP & Unknown & Unknown & Unknown & 2 Billion \\
Polish RoBERTa (large) & 30k & 50k & ~135GB & **10-20 Billion \\
\bottomrule
\end{tabular}
\setlength{\abovecaptionskip}{4pt}
\caption{Model Comparison with Batch Size, Update Steps, Corpus Size, and Tokens Seen.}
\footnotesize{**This presents an estimated range of token numbers derived from the cited datasets, inferred due to the lack of explicit mention in the associated repositories or papers.}
\label{table:model_comparison_2}
\end{table}

\section{Power usage, costs and carbon offset}
\textbf{RQ 3: What are the estimated costs, time requirements, and energy consumption involved in building a model like Curie-7B-v1?} \label{answer-rq3}
\\
The training of the model was carried out using a cloud provider. It took 106 GPU hours and incurred a cost of \$85 for the first stage of the LAPT. Additionally, approximately \$50 was spent to train and fine-tune hyperparameters of nine different classifiers, requiring around 60 GPU hours in a cloud setup. The approximated power consumption of the server for the whole training time can be calculated in the following way.
\begin{equation}
 450W \times 166h = 74.7 kWh
\end{equation}
The estimated server power consumption of 74.7 kWh will be used to approximate carbon offset. The carbon emission was calculated using approximated carbon produced based on the local power grid as follows:
\begin{equation}
    74.7kWh\ \times \sim 0.61 \, \ kg\ eq. \, \ch{CO2} / kWh \approx 45.57 \, \ kg\ eq. \, \ch{CO2}
\end{equation}
This calculation underscores the environmental efficiency of the proposed solution. There is no necessity to develop a foundational model, which often demands extensive training on 8, 16, 32, or even hundreds of GPUs over several days for marginally improved performance. Such an effort has already been undertaken by the Mistral-AI team during the pre-training stage. In the case of the classifiers, the inference speed was remarkably fast on both an 80-watt CPU and a 300-watt GPU. Employing techniques such as pruning or quantization could further enhance environmental friendliness, reducing memory requirements and improving efficiency.
\section{Conclusions}
In this paper, we introduce Language Adaptive Pre-training (LAPT) applied in the Curie-7B-v1 model, a decoder-only architecture inspired by clinical ML research. The LAPT approach demonstrates that the Curie-7B-v1 model matches foundational Polish models. On eight downstream tasks, it achieved an average score of 89.35\% compared to the top model's 90.7\%. Curie-7B-v1 achieved this score but with markedly fewer data utilizing just 2-3\% of the dataset size. Unlike compared traditional encoder-decoder models limited to predicting masked tokens, Curie-7B-v1 exhibits versatility in generating high-quality Polish text. This adaptability allows adapting it to various problems, including classification, regression, and text generation. 
The integration of 2-bit quantization \cite{chee2024quip} and pruning methods into the adaptation of LLMs for low-resource languages could be a valuable area for future research. These strategies promise to improve the efficiency and accessibility of language models. This model fills a crucial gap by providing an open-source Polish LLM, laying the groundwork for developing modern, efficient business solutions.

\section{Acknowledgements}
We acknowledge the financial support from Apostroph Group and express appreciation for Dr. Tomer Jack Barnea, Head of ICT in Apostroph Group, for his support in my AI research. Their assistance provided the necessary resources and expertise to overcome the challenges faced and the development of this project.

\printbibliography

\end{document}